Title
- AI-Driven Reinvention of Hydrological Modeling for Accurate Predictions and Interpretation to Transform Earth System Modeling

Authors

Cuihui Xia,[1] Lei Yue,[2] Deliang Chen†[3,4], Yuyang Li[5], Hongqiang Yang[6], Ancheng Xue[7], Zhiqiang Li[2], Qing He[8], Guoqing Zhang[1], Dambaru Ballab Kattel[1], Lei Lei[9], Ming Zhou[9]

Affiliations
1. Institute of Tibetan Plateau Research, Chinese Academy of Sciences, Beijing, China
2. China Electric Power Research Institute, Beijing, China
3. Department of Earth System Science, Tsinghua University, Beijing, China
4. Department of Earth Sciences, University of Gothenburg, Gothenburg, Sweden
5. National Astronomical Observatories of the Chinese Academy of Sciences, Beijing, China
6. The South China Sea Institute of Oceanology, Chinese Academy of Sciences, Guangzhou, China
7. North China Electric Power University, Beijing, China
8. Global Energy Interconnection Development and Cooperation Organization, Beijing, China
9. Alibaba Cloud, Hangzhou, China

†Corresponding author：Deliang Chen
**deliangchen@tsinghua.edu.cn**

CX and LY contribute equally to the work

Abstract
Traditional equation-driven hydrological models often struggle to accurately predict streamflow in challenging Earth systems like the Tibetan Plateau, while hybrid and existing algorithm-driven models face difficulties in interpreting hydrological behaviors. This study introduces HydroTrace, an algorithm-driven, data-agnostic model that substantially outperforms these approaches, achieving a Nash-Sutcliffe Efficiency of 98% and demonstrating strong generalization on unseen data. Moreover, HydroTrace leverages advanced attention mechanisms to capture spatial-temporal variations and feature-specific impacts, enabling the quantification and spatial resolution of streamflow partitioning as well as the interpretation of hydrological behaviors such as glacier-snow-streamflow interactions and monsoon dynamics. Additionally, we developed a large language model (LLM)-based application that allows users to easily understand and apply HydroTrace's insights for practical purposes. These advancements position HydroTrace as a transformative tool in hydrological and broader Earth system modeling, offering enhanced prediction accuracy and interpretability.

**Teaser**

HydroTrace is an interpretable AI-driven model for streamflow predictions, offering detailed and easy-to-understand insights via a multilingual app.

**MAIN TEXT**

**Introduction**

Earth System Science (ESS) encompasses the study of the interacting physical, chemical and biological processes between the atmosphere, hydrosphere, biosphere, cryosphere, lithosphere, and anthroposphere. These systems, while distinct, are interconnected and influence each other in ways that govern the behavior of the Earth as a whole. Earth System Modeling (ESM) serves as a critical tool in understanding these interactions, providing a framework for simulating the dynamics of Earth's processes across multiple scales. Models in ESS are typically built on physical, chemical, and biological principles that capture the patterns of these interactions, enabling predictions about climate change, ecosystem responses, and resource management. However, the challenge lies in the complexity of accurately representing these interactions, especially under changing environmental conditions such as climate variability and human-induced alterations(*1–4*).

Hydrological modeling is a cornerstone of earth system process modeling, crucial for understanding and managing the movement, distribution, and quality of water. Hydrological modeling has advanced remarkably, leveraging increasingly complex models and diverse data sources to simulate water-related processes. Developments include the adoption of distributed models, physically based approaches, and machine learning techniques, which enhance spatial-temporal accuracy and enable the integration of diverse hydrological variables(*5–7*).

However, challenges remain. Traditional deterministic models often oversimplify hydrological systems, struggling with nonlinear processes, data sparsity, and heterogeneity, particularly in complex terrains like the Tibetan Plateau (*5*, *6*, *8*). While machine learning improves predictions, its integration into hydrological modeling is constrained by issues of interpretability(*7*, *9*, *10*). Similar challenges exist in broader Earth system process modeling, where hybrid approaches are increasingly sought to address the limitations of equation-driven models in capturing nonlinear and complex systems, and of algorithm-driven models in providing meaningful interpretability (*11–13*).

While hybrid models offer incremental improvements by balancing predictive accuracy, generalizability, and interpretability (*9*, *10*, *12*, *14*), there are always compromises among the three (*15*, *16*). This is because equation-driven and algorithm-driven approaches represent fundamentally different paradigms(*17*, *18*). Equation-driven models rely on predefined physical assumptions to simulate processes, while algorithm-driven models extract patterns and relationships directly from data, enabling them to adapt to complex, nonlinear systems (*7*, *15*, *17*). Merging the two would result in a scenario where one approach is dominant and the other supplementary, necessitating compromises in either predictive accuracy, generalizability, or interpretability.

Algorithm-driven approaches align more closely with the principles of science, offering universal applicability akin to mathematical theorems and the promise of robust, data-driven predictions that are testable and falsifiable(*19*). In contrast, physical equations often require location-specific adjustments, limiting their scalability, generalizability, and universal applicability(*5*, *16*). This reliance on context-specific formulations underscores their empirical foundation, as no single equation can universally represent complex Earth systems. These limitations highlight the need for a paradigm shift toward algorithm-driven models, which can adapt to diverse systems without relying on rigid local assumptions.

However, as **Fig. 1** shows, algorithm-driven models are frequently relegated to roles such as data assimilation and parameter optimization rather than defining model structures in hybrid hydrological modeling(*9*, *16*). In the rare cases where models are built in an algorithm-driven manner(*10*), interpretability often reverts to the equation-driven deterministic approach. This is because the algorithms used in hydrological modeling are primarily "black box" methods such as artificial neural networks (ANNs)(*20*), convolutional neural networks (CNNs), recurrent neural networks (RNNs), and their variant long short-term memory networks (LSTMs)(*10*). Extracting physical insights from these "black boxes" is challenging; therefore, interpretability considerations should be incorporated during the algorithm design(*21*). To fully leverage the potential of algorithm-driven models, there is a need for innovation in algorithms that can advance all three aspects—predictive accuracy, generalizability, and interpretability—without compromise. Since predictive accuracy and generalizability are known strengths of the algorithm-driven approach, this suggests that the key scientific question in hydrological modeling is not how to enhance equation-driven models with machine learning, but how to make algorithm-driven models interpretable.

Here we propose a paradigm shift for hydrological modeling and the broader earth system modeling: the development of robust, generalizable algorithms capable of delivering exceptional predictive accuracy, interpretability, and adaptability across diverse systems. HydroTrace exemplifies this vision. **Fig. 1** demonstrates the evolution from manual calibration and region-specific adjustments in traditional models, through the integrated and semi-automatic approaches of hybrid models, to the fully automated and data-agnostic calibration processes of algorithm-driven models, underscoring the enhanced flexibility and interpretability of HydroTrace achieved through attention-based algorithm design. Developed without pre-analyzing input data patterns or their physical relationships, HydroTrace leverages two key attention-based(*22*) algorithms under the hypothesis that river streamflow is shaped by variations of land surface features across time and space. One algorithm dynamically focuses on specific regions at each timestep, while the other identifies key features within each spatial grid. These specially designed mechanisms have proven effective in enabling HydroTrace to make accurate predictions and generate interpretations consistent with observations. We believe these attention-based algorithms offer a transformative approach to hydrological modeling and the broader earth system modeling, which fundamentally revolves around feature variations across time and space.

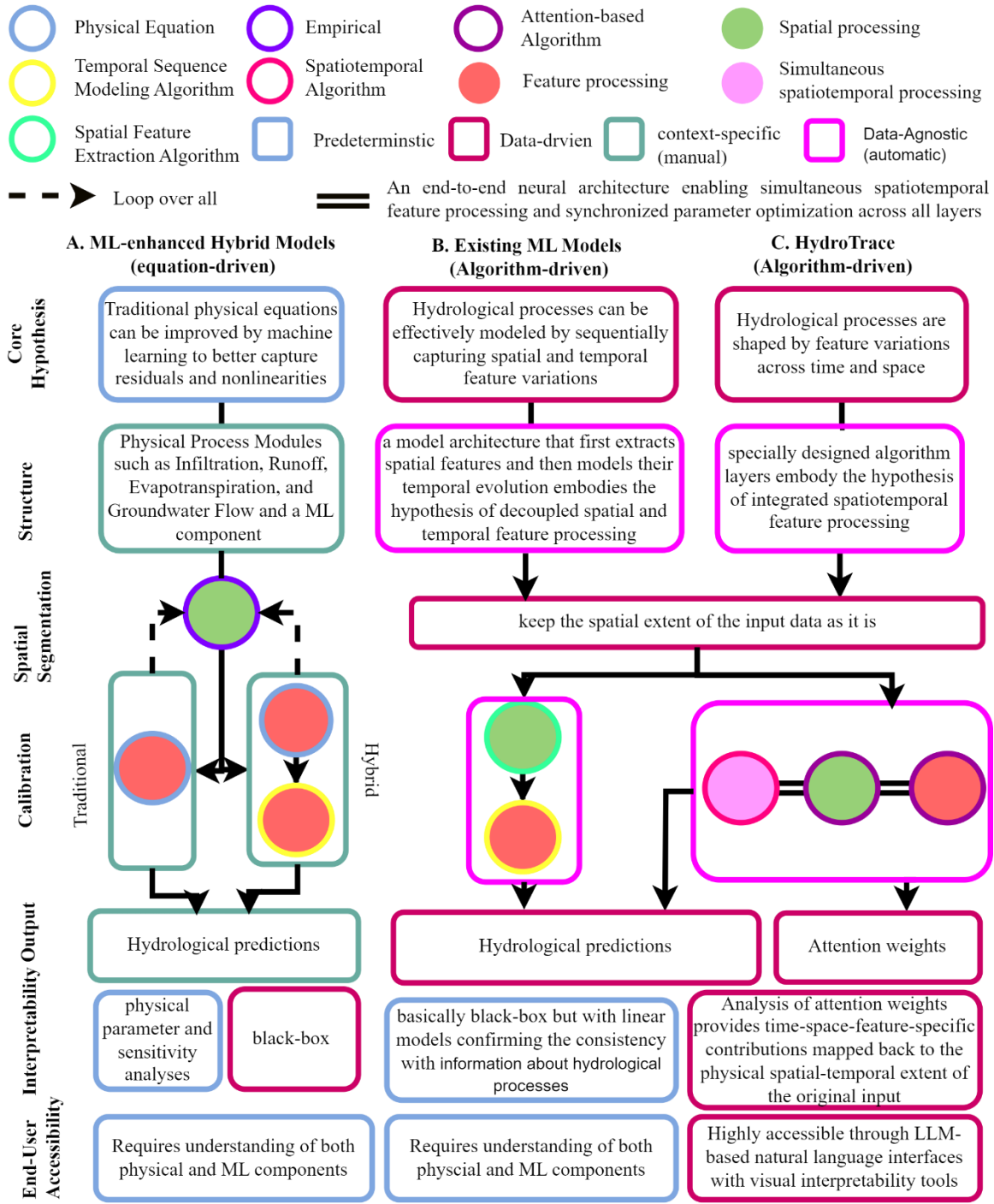

**Figure 1: Comparative structure and modeling processes of equation-driven and algorithm-driven hydrological models.** Panel A illustrates machine learning (ML)-enhanced hybrid models, highlighting their deterministic, process-based modules alongside manual and region-specific calibration and parameterization methods. Panel B depicts existing ML Models, which sequentially capturing spatial and temporal feature variations with limited interpretability. Panel C showcases the HydroTrace Model, featuring spatiotemporal algorithm and specially designed attention-based algorithm layers, which enable processing spatiotemporal feature automatically and simultaneously, mapping attention weights back to the physical extent of input data exactly, and improving interpretability with an LLM-based interface for easy user access.

# Results

We present results to demonstrate the performance and capabilities of HydroTrace across three key aspects. First, we evaluate its predictive accuracy and generalizability in one of the most complex Earth systems on the Tibetan Plateau, showcasing its ability to handle both in-sample and out-of-sample data effectively. Next, we highlight HydroTrace's interpretability by using it to address key hydrological questions typically explored through equation-based models. Finally, we present a case study illustrating an application of HydroTrace in hydropower management, enabling practical insights that were previously unattainable with traditional approaches.

## Predictive Accuracy and Generalizability of HydroTrace Compared to Equation-driven Models

Our study focuses on a section of the upper Brahmaputra Basin, specifically the middle reach of the Yarlung Zangpo River and its tributary, the Lhasa River, on the Tibetan Plateau. This region is characterized by data scarcity, active hydrological processes, and a complex earth system(*8*, *23*). The study area **(Fig. 2)** encompasses the glacier-rich Nyainqêntanglha Range, a critical climatic and hydrological divide influenced by westerlies from the northwest and monsoons from the southeast (*24*, *25*). This unique geographical setting serves as an ideal natural laboratory for evaluating HydroTrace's effectiveness in addressing interactions among atmospheric circulations, the cryosphere, and hydrology.

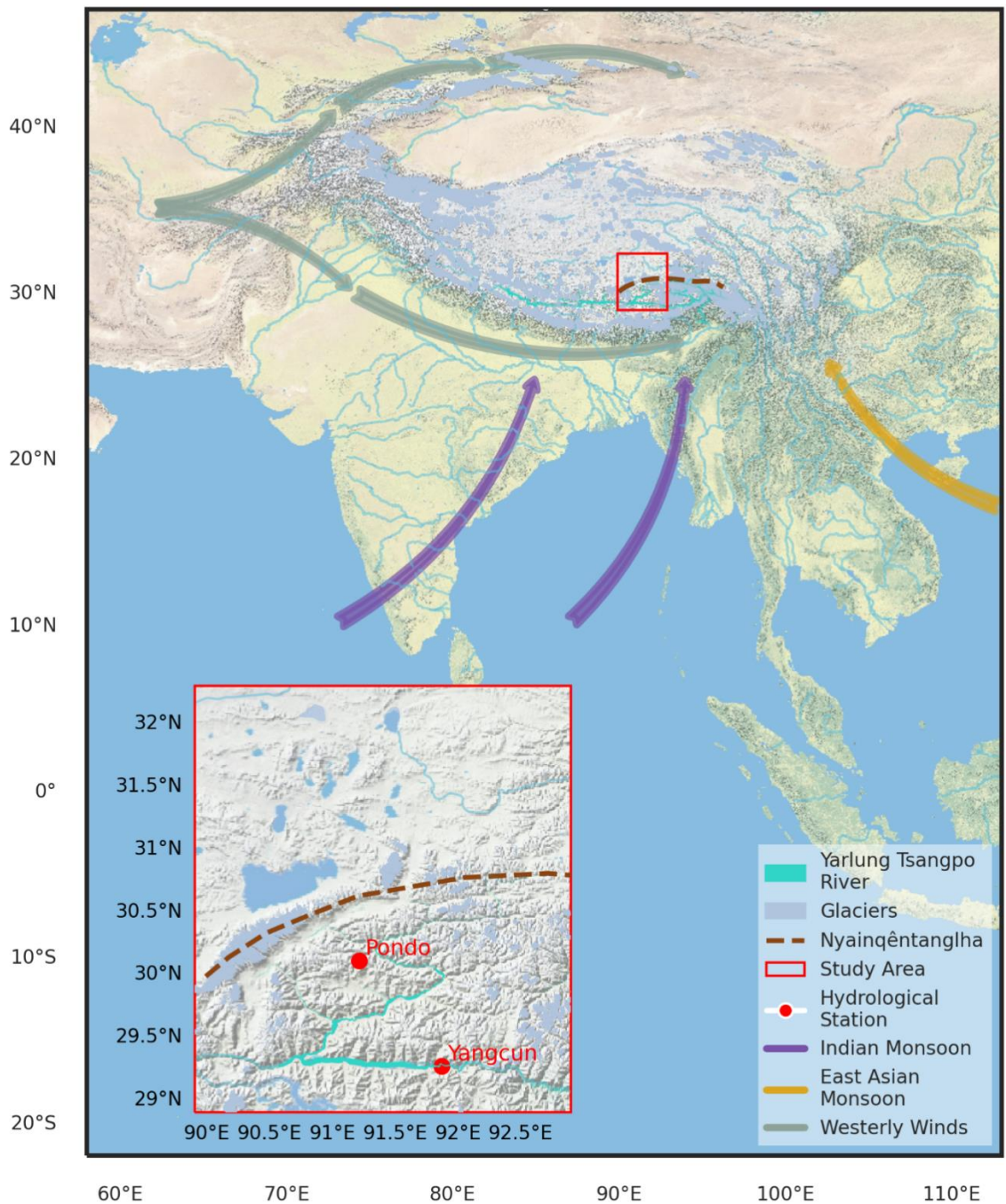

**Fig. 2. The study area hosting complex earth system processes.** The study area represents a region characterized by diverse land surface features and intensive atmosphere-land interactions. The Nyainqêntanglha range, prominently outlined on the Esri basemap, acts as a critical boundary influencing regional processes. The map highlights glaciers, the Yarlung Tsangpo River, and key atmospheric circulations, including the Indian Monsoon, East Asian Monsoon, and Westerly Winds. Hydrological stations are marked to indicate locations near which HydroTrace is calibrated using data from operational companies.

As shown in **Fig. 2**, our study area was not defined by basin boundaries. HydroTrace, as a data-agnostic model, does not require a basin-specific framework to function. The basin concept is applied here solely to enhance understanding and provide a familiar framework for peers, ensuring clarity in interpretation. We trained, or in hydrological terms, calibrated HydroTrace at two distinct hydrological sites, Yangcun and Pondo. Notably, unlike previous models that rely on hydrological observation stations at these locations, HydroTrace uses data recorded by companies operating near the stations. Due to data licensing restrictions requiring anonymity, all streamflow values in **Fig. 3** have been normalized to maintain confidentiality. **Fig. 3** also illustrates HydroTrace's convergence during both calibration and validation phases at both sites. Its consistently high performance at these two distinct locations using the same algorithmic structure demonstrates its robustness as a data-agnostic hydrological model.

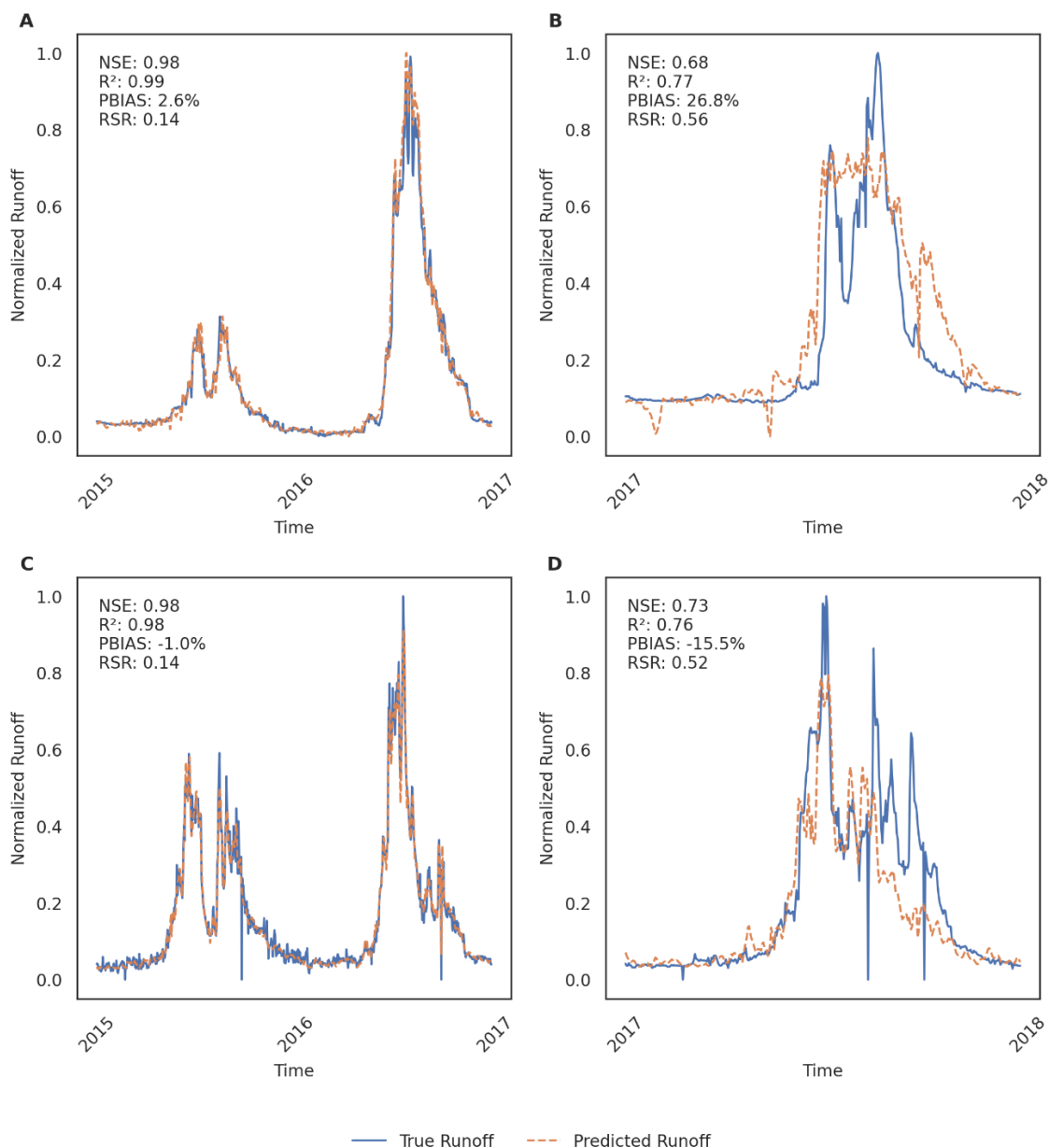

**Figure 3. Performance of HydroTrace for streamflow prediction across different locations.** (A) Calibrated performance (2015–2016) for the Yangcun site. (B) Validation performance (2017) for the Yangcun site. (C) Calibrated performance (2015–2016) for the Pondo site. (D) Validation performance (2017) for the Pondo site. Solid lines represent the true normalized streamflow, and dashed lines represent the predicted normalized

streamflow. Each panel annotates performance metrics: Nash-Sutcliffe Efficiency (NSE), Coefficient of Determination (R²), Percent Bias (PBIAS), and RMSE-Observation Standard Deviation Ratio (RSR). Normalization was applied to anonymize streamflow values, ensuring they fall within the range [0.0, 1.0], as required by our data license to preserve data anonymity.

HydroTrace demonstrated exceptional performance across various hydrological metrics during both calibration and validation phases **(Fig. 3)**. The evaluation criteria for these metrics align with thresholds established in the literature(*26*, *27*), allowing for clear categorization of performance levels such as "Very Good" and "Good." The results underscore HydroTrace's robustness and potential to address complex hydrological challenges effectively.

**Nash-Sutcliffe Efficiency (NSE):** During calibration, HydroTrace achieved NSE values of 0.98 at both Yangcun (Fig. 3, A) and Pondo(Fig. 3, C), firmly within the "Very Good" range(*27*). Validation results showed strong performance, with an NSE of 0.68 at Yangcun(Fig. 3, B), categorized as "Satisfactory," and 0.73 at Pondo(Fig. 3, D), categorized as "Good"(*27*). These findings highlight HydroTrace's ability to reliably replicate observed streamflow across varying hydrological contexts and timeframes.

**Percent Bias (PBIAS):** HydroTrace displayed minimal calibration bias, with PBIAS values of 2.6% at Yangcun(Fig. 3, A) and -1.0% at Pondo(Fig. 3, C), both classified as "Very Good."(*27*) During validation, PBIAS increased to 26.8% at Yangcun, reflecting overestimation, while PBIAS at Pondo was -15.5%, falling within the "Satisfactory" range. The disparity in performance is likely attributable to the distinct hydrological characteristics of the sites. Yangcun, positioned within the mainstream and middle range of the Yarlung Tsangpo River, is subject to greater complexity from tributary inflows and dynamic flow interactions. This complexity can increase predictive uncertainty when training data is limited. Conversely, Pondo's upstream location as a single-branch system results in simpler hydrological dynamics, contributing to better model performance at this site. Expanding the training data to include longer time series and a broader spatial extent would enable HydroTrace to better capture the diverse hydrological influences at Yangcun, improving its performance and reducing bias.

**Root Mean Square Error (RSR):** HydroTrace recorded an RSR of 0.14 during calibration at both sites, reflecting "Very Good" accuracy. Validation RSR values were 0.56 at Yangcun(Fig. 3, B) and 0.52 at Pondo(Fig. 3, D), both categorized as "Good." (*26*, *27*) These consistent results across metrics emphasize HydroTrace's reliability and its ability to maintain predictive accuracy across varying conditions.

**Coefficient of Determination (R²):** Calibration R² values of 0.99 at Yangcun(Fig. 3, A) and 0.98 at Pondo(Fig. 3, C) underscore HydroTrace's ability to capture variability in observed data with "Very Good" accuracy(*27*). Validation R² values of 0.77 at Yangcun(Fig. 3, B) and 0.76 at Pondo(Fig. 3, D), categorized as "Good,"(*27*) further confirm the model's effectiveness in explaining observed hydrological variability.

Overall, the consistently outstanding performance of HydroTrace across both Yangcun and Pondo highlights its predictive superiority, making it a more reliable and robust choice for hydrological modeling. An added value of HydroTrace is its lower time, labor, and financial costs. For two years of daily data, HydroTrace requires only about 5 hours to complete

training—or calibration, in hydrological terms—automatically on a commonly available and affordable A10 GPU. This efficiency makes HydroTrace much faster than equation-driven models and reduces the barriers for end users to apply it in real-world scenarios.

**Interpretability of Hydrological Processes Using HydroTrace**

The Tibetan Plateau's complex climate-cryosphere-hydrology system poses significant challenges for hydrological modeling. Previous studies(*8, 23*) have identified three key issues crucial for regional water studies: integrating cryosphere processes into hydrological models, quantifying and spatially resolving streamflow partitioning, and understanding the interplay between monsoonal and westerly influences on streamflow dynamics. Equation-driven hydrological models, constrained by limited data and context-specific formulations, have struggled to address these challenges.

To demonstrate HydroTrace's interpretive capabilities, we trained the model using daily data from the Yangcun site spanning 2015–2017, utilizing the full dataset rather than separating it into calibration and validation periods. This approach prioritizes interpretability over predictive testing, enabling the model to fully leverage its temporal-spatial and feature-wise attention mechanisms to uncover patterns and insights across the entire data range. HydroTrace consistently achieved high performance with an NSE of 0.98 (see Fig. S1). The attention weights derived from this interpretive model provide a detailed assessment of the impact of each input feature, grid cell, and time step on streamflow estimations, allowing us to comprehensively address the region's three key hydrological challenges (see Materials and Methods for details)

*Cryosphere-hydrology interaction analysis*

We visualized the attention weights of glacier-, snow-, and precipitation-related features over the months in **Fig. 4** to illustrate the complicated interactions between the cryosphere and hydrology at the Yangcun site.

During the winter months, particularly in December and January, the attention weights for Glacial Runoff (Monthly) and Snow Depth peak at approximately 0.0845 (Fig. 4, A) and 0.0063(Fig. 4, B), respectively. This indicates that both glacial meltwater and substantial snow accumulation play key roles in sustaining streamflow during these periods. The high attention weight for glacial runoff in December underscores the significant contribution of glacial meltwater when precipitation is relatively low. Meanwhile, the growing attention weight for snow depth in January highlights the accumulation of snowpack, which acts as a critical reservoir for future meltwater contributions. The presence of deep snowpacks enhances surface albedo, as evidenced by the high attention weights for Snow Age in October (0.0128) and January (0.0116), reflecting the influence of freshly fallen and aging snow on melting dynamics (Fig. 4, B). This combination suggests glacial runoff and substantial snow depth during colder months are key to maintaining streamflow.

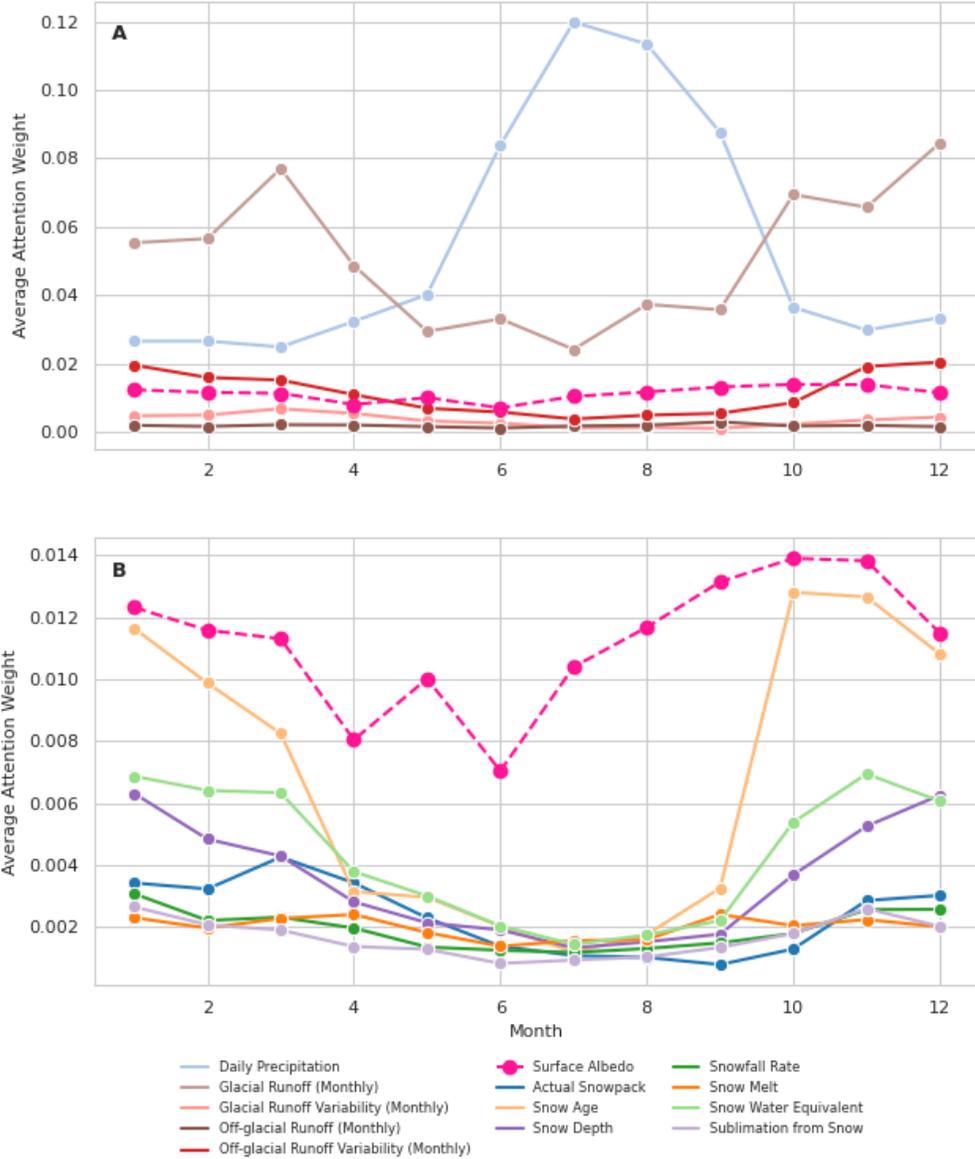

**Figure 4. Monthly variation in attention weights for glacier- and snow-related features.** (A) Attention weights of glacier-related features: Glacial Streamflow (Monthly), Glacial Streamflow Variability (Monthly), Off-glacial Streamflow (Monthly), Off-glacial Streamflow Variability (Monthly), and Surface Albedo. (B) Attention weights of snow-related features: Snowfall Rate, Snow Melt, Snow Water Equivalent, Snow Depth, Snow Age, Sublimation from Snow, and Actual Snowpack, with Surface Albedo included. Both subplots share the same x-axis, representing 12 months. Lines represent the monthly average attention weights for each distinctively colored feature.

During the pre-monsoon months March and April, there is a notable increase in the attention weight for Snow Melt, reaching a peak of approximately 0.0024 in April (Fig. 4, B). This surge signifies the critical period when rising temperatures initiate significant snowmelt, directly enhancing streamflow. Simultaneously, Glacial Runoff (Monthly) shows higher attention weights, particularly in March (0.0770), indicating that glacier ice continues to contribute to streamflow as the snowpack begins to diminish (Fig. 4, A). The interplay between melting snow and exposed darker glacier ice accelerates melt rates, driven by decreased surface albedo as snow cover reduces. This transition is further supported by the declining attention weights for Snow Depth and Snow Water Equivalent (SWE), which reflect the gradual depletion of the snowpack (Fig. 4, B).

With the onset of monsoon season, particularly in June, July, and August, the hydrological dynamics shift dramatically. The attention weight for "Daily Precipitation" peaks in July (0.119869) and August (0.113415) (Fig. 4, A), highlighting the overwhelming influence of monsoon rainfall on streamflow. This substantial increase in precipitation overshadows the relative contributions from glacial runoff and snowmelt, as evidenced by the sharp decline in Glacial Runoff (Monthly) attention weights, dropping to 0.0242 in July (Fig. 4, A).

During the post-monsoon months of October and November, the system begins to reset for the next seasonal cycle. In October and November, Surface Albedo shows higher attention weights (0.013908 and 0.013828, respectively), indicating the accumulation of fresh snow that increases surface reflectivity. This high albedo reduces the absorption of solar radiation, slowing down melt rates and preserving the snowpack and underlying glacier ice. The peak in Snow Age in October (0.0128) further supports this (Fig. 4, B), as newly fallen snow with high albedo and low density insulates the glaciers, limiting immediate melting.

Throughout the year, a key indicator of the cryosphere, Surface Albedo, plays a continuous yet modulating role in streamflow dynamics. Its peaks in October and November reflect periods of high reflectivity due to fresh snow cover, which collectively influence melting rates by controlling the amount of solar energy absorbed by the glacier and snow surfaces (Fig. 4).

With attention weights extracted from HydroTrace, we obtained a detailed understanding of how the cryosphere influence hydrology at the studied site. The streamflow at the Yangcun site is characterized by a seasonal transition driven by the interplay of glacial runoff, snow dynamics, and precipitation. During the winter and pre-monsoon months, glaciers and snowpacks are essential in sustaining streamflow, with glacial runoff and snowmelt providing critical water inputs when precipitation is low. The dynamics of surface albedo and snow properties, such as age and depth, modulate these contributions by affecting melting rates and preserving snowpack integrity. With the onset of monsoon season, the dominance of direct rainfall reduces the relative impact of cryospheric features on streamflow, shifting the hydrological balance towards precipitation-driven dynamics.

Notably, the comprehensive and temporally consistent analysis of feature contributions provided by HydroTrace cannot be matched by traditional equation-driven models using sensitivity tests, or by existing algorithm-driven hydrological methods that rely on model-agnostic interpretation tools such as SHapley Additive exPlanations (SHAP)(*28*). This highlights HydroTrace's enhanced interpretability and effectiveness in hydrological studies.

*Quantifying and spatially resolving streamflow partitioning*

In terms of quantifying and spatially resolving streamflow partitioning, we designed the attention weights to be scaled between 0 and 1 (see Materials and Methods for details) to facilitate their interpretation as streamflow partitioning metrics. This scaling enabled **Fig. 5**, which lists the top five contributors to streamflow by season along with their corresponding spatial distributions. For a detailed full list of feature contributions, see Fig. S2.

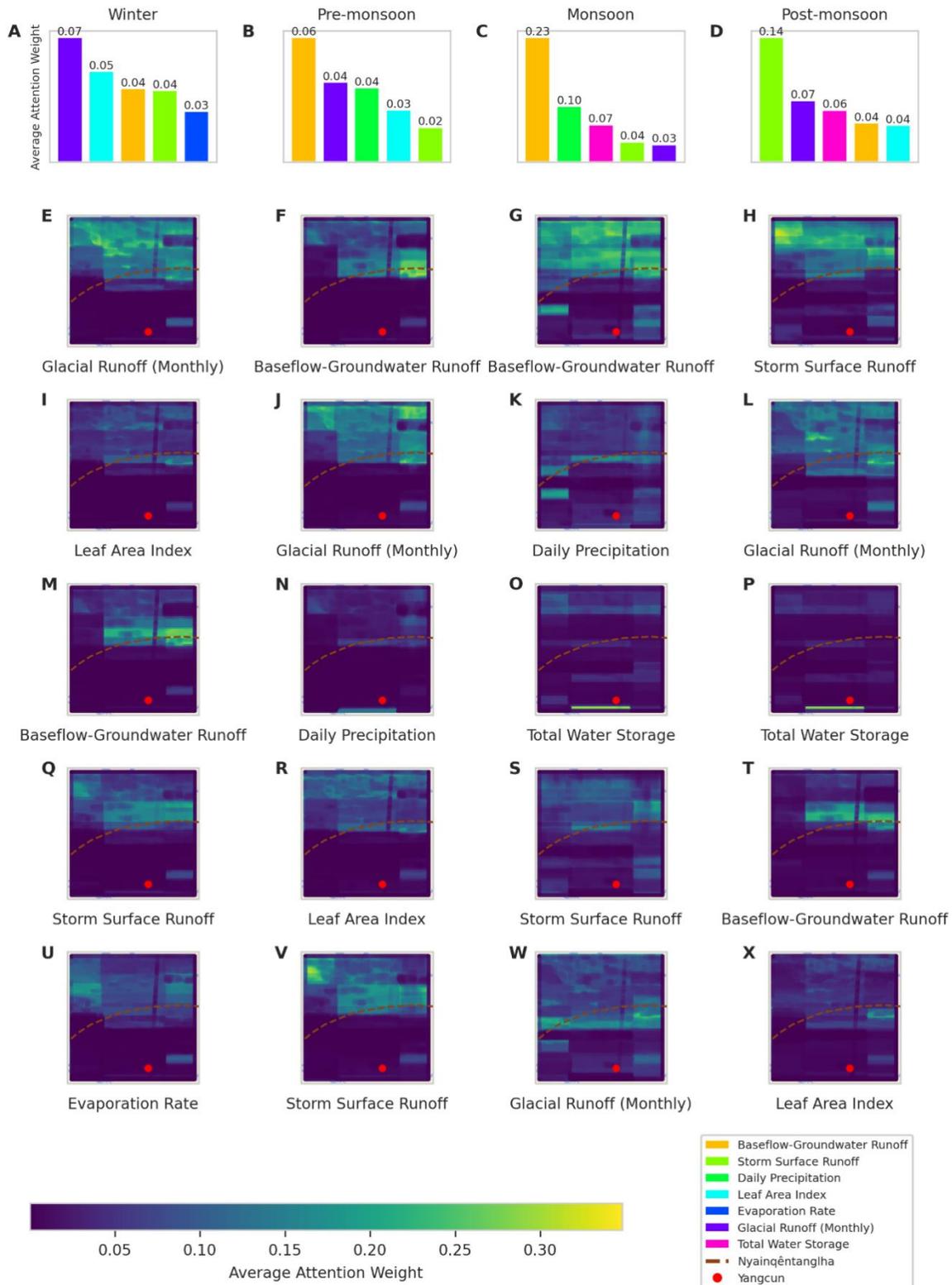

**Figure 5. Top 5 Contributing Features by Season and Their Spatial Attention Distributions.** The first row (A–D) presents bar charts illustrating the five features with the highest average attention weights for each of the four defined seasons: Winter, Pre-monsoon, Monsoon, and Post-monsoon. Seasons are divided as follows: Winter (December–February), Pre-monsoon (March–May), Monsoon (June–September), and Post-monsoon (October–November). Each feature is assigned a distinct color, consistent across all seasons, with corresponding average attention weight values annotated above each bar. The subsequent five rows (E–P) display maps depicting the spatial distribution of these top features for each season. The color gradient on the maps represents the magnitude of average attention weights, as indicated by the color bar below.

Each season exhibits a distinct ranking of the highest contributing features to streamflow at Yangcun. During the Winter season, Glacier Streamflow is the predominant contributor, accounting for 7% of the total streamflow, followed by Leaf Area Index (5%), Baseflow and Groundwater Streamflow (4%), Storm Surface Streamflow (4%), and Evaporation Rate (3%) (Fig. 5, A). As the Pre-monsoon season arrives, the contribution of Glacier Streamflow decreases to 4%, while Baseflow and Groundwater Streamflow ascend to lead with a 6% contribution. Daily Precipitation and Leaf Area Index each contribute 4%, and Storm Surface Streamflow accounts for 2% (Fig. 5, B). In the Monsoon season, Baseflow and Groundwater Streamflow dominate significantly, contributing 23% to streamflow. This is followed by Daily Precipitation (10%), Total Water Storage (7%), Storm Surface Streamflow (4%), and Glacier Streamflow (3%) (Fig. 5, C). During the post-monsoon season, Storm Surface Streamflow becomes the leading contributor at 14%, followed by Glacier Streamflow (7%), Total Water Storage (6%), Baseflow and Groundwater Streamflow (4%), and Leaf Area Index (4%) (Fig. 5, D).

Glacier streamflow consistently exerts a strong influence on streamflow at Yangcun throughout the year; however, its key zones with highest influence shift spatially from the northwest to the southeast across seasons (Fig. 5, E, J, W, and L). This spatial movement reflects the transition in climatic dominance from the westerlies to the monsoon, as the primary areas of glacier streamflow impact shift from the northwest to the southeast of the Nyainqêntanglha Range—a critical climatic divide between westerly and monsoonal influences. This indicates that while glacier streamflow impacts are present year-round, they originate from different circulation-dominant zones and are governed by distinct climatic controls.

Baseflow and Groundwater Streamflow remain strong, year-round contributors to streamflow at Yangcun. Their most influential zones are consistently centered on the southeastern part of the Nyainqêntanglha during Winter, Pre-monsoon, and post-monsoon seasons (Fig. 5, M, F and T). However, during the Monsoon season, the influential zone expands to encompass the entire region above the Nyainqêntanglha, indicating a monsoon-driven enhancement of baseflow dynamics (Fig. 5, G). This seasonal expansion suggests that increased precipitation and enhanced infiltration rates during the Monsoon season augment Baseflow and Groundwater Streamflow, sustaining groundwater levels and enhancing their contribution to streamflow.

Storm Surface Streamflow exerts a strong influence on streamflow across all seasons, with its spatial dynamics shifting in response to seasonal climatic patterns. During the monsoon season, its influence is concentrated in areas where the Nyainqêntanglha Range fragments, and westerly and monsoonal circulations converge, amplifying streamflow through intense precipitation and storm activity (Fig. 5, S). In the pre-monsoon and post-monsoon seasons, its highest-impact zone shifts to the glacial lake-intensive northwestern corner of the study area (Fig. 5, V and H), where "hidden glacier melt"—glaciers melting into glacial lakes(*29*)—could modulates streamflow by storing meltwater and releasing it during precipitation events. During winter, the key influential zone of Storm Surface Streamflow spans multiple regions, including the southeast of the Nyainqêntanglha Range, the northwestern corner, and the areas in between (Fig. 5, Q). This spatial distribution reflects a combination of storm-induced streamflow together with glacial lakes acting as reservoirs to sustain streamflow even in colder, drier months.

The Leaf Area Index (LAI) strongly influences streamflow in all seasons except the Monsoon, with its impact concentrated in the relatively stable northern upstream area, particularly where westerly and monsoonal influences converge and glaciers are present (Fig. 5, I, R, and X). This indicates that vegetation plays a critical role in intercepting precipitation, regulating evapotranspiration, and modulating surface streamflow. In glacier-adjacent regions, vegetation may further affect hydrological processes by stabilizing soil and retaining moisture from glacier meltwater, thereby shaping the timing and volume of streamflow.

Evaporation Rate ranks among the top five contributors only during the Winter season, with its influential zones centered in lake-rich areas (Fig. 5, U). This indicates that in precipitation-scarce winter conditions, lakes contribute significantly to streamflow dynamics at Yangcun by affecting local evaporation rates.

Surprisingly, Daily Precipitation ranks among the top five contributors only during the Pre-monsoon and Monsoon seasons. This can be attributed to the highly seasonal nature of precipitation on the Tibetan Plateau, where most rainfall occurs during the Monsoon, while winter and Post-monsoon precipitation is minimal and often stored as snow. Additionally, data limitations, including sparse observations and complex topography, make it challenging to accurately simulate precipitation across seasons. During the Pre-monsoon season, its highest influence is concentrated on the southern bank of the Yarlung Zangpo River near the Yangcun site (Fig. 5, N), while in the Monsoon season, it shifts to areas where glaciers on the Nyainqêntanglha Range meet branches of the Yarlung Zangpo River (Fig. 5, K). This spatial shift suggests that streamflow at Yangcun is primarily driven by the interaction between precipitation and glacial melt, with rain-on-glacier processes playing a key role during the Monsoon season.

Total Water Storage appears among the top five contributors during the Monsoon and Post-monsoon seasons, with its key influential zones centered on the southern bank of the Yarlung Zangpo River (Fig. 5, O and P). This highlights the importance of water storage mechanisms in buffering streamflow responses to intense precipitation events, thereby influencing overall hydrological behavior.

Again, such detailed and spatially resolved analysis of streamflow partitioning is not achievable with traditional equation-driven models or existing algorithm-driven models, further underscoring the superior interpretability and effectiveness of HydroTrace in hydrological studies.

*Westerly-Monsoon Dynamics*

In analyzing the dynamics of westerly-monsoon interactions, we conducted a spatial analysis of attention weights by visualizing the geographic distribution of the top 20% of locations with the highest average attention weights for each respective season, as illustrated in **Fig. 6**.

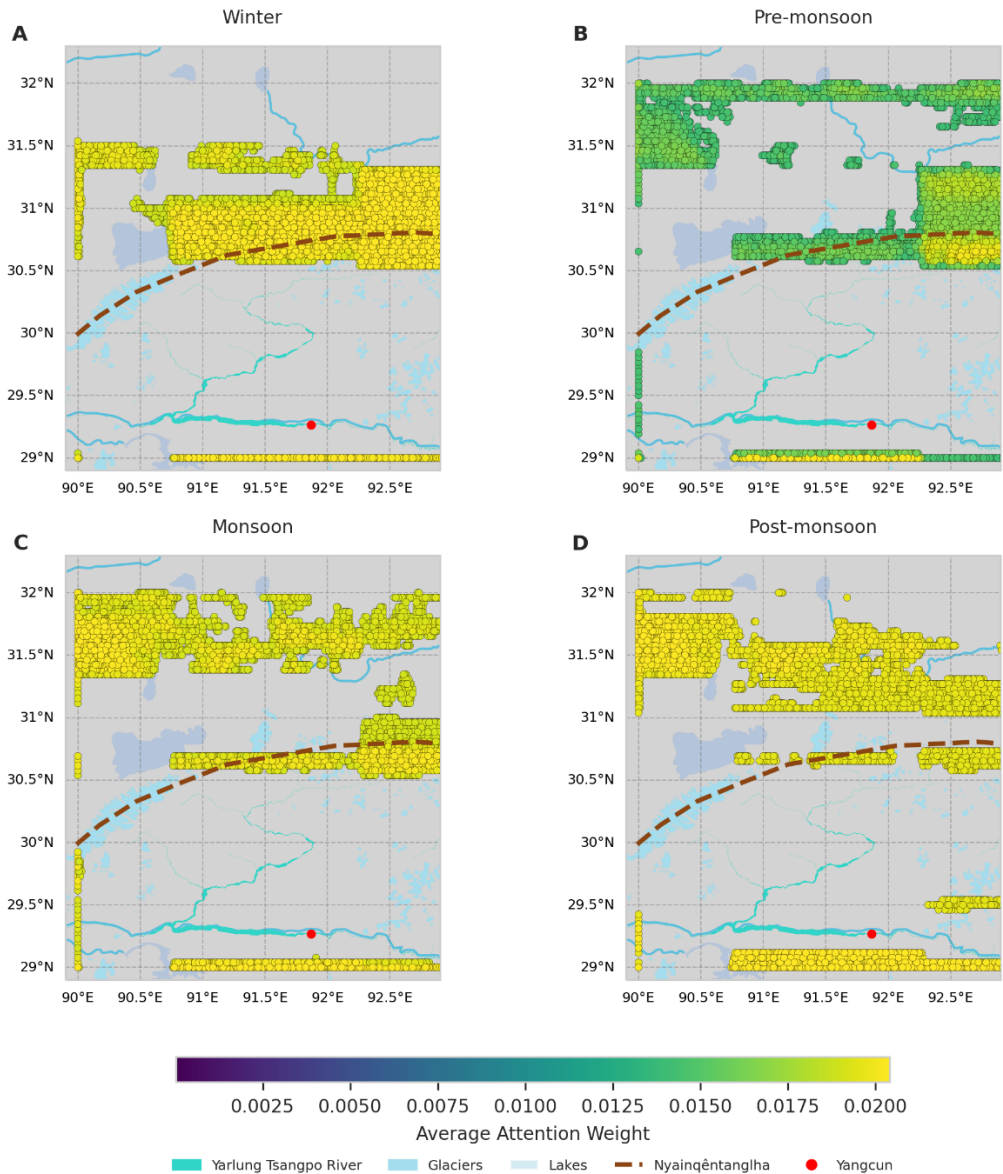

**Figure 6. Spatial Distribution of the Most Influential Locations by Season.** Panels A–D display the geographic distribution of the top 20% locations with the highest average attention weights for each respective season: Winter (A), Pre-monsoon (B), Monsoon (C), and Post-monsoon (D). Seasons are divided as follows: Winter (December–February), Pre-monsoon (March–May), Monsoon (June–September), and Post-monsoon (October–November). Each subplot overlays the significant locations on a consistent basemap that highlights key geographic features, including rivers, glaciers, and mountain ranges. The color intensity of the heatmap represents the magnitude of the average attention weight, utilizing a unified color scale to facilitate comparison across all seasons.

During the winter season, when the monsoon's influence is at its weakest, the primary influence zone is centered in the region where the mountain ranges are fragmented (Fig. 6, A). This fragmentation creates an interaction zone where the westerlies meet the monsoon circulation, establishing what can be considered the default state of westerly-monsoon interaction. The concentration of high attention weights in this area underscores the significant role that the interaction between these two circulation systems plays in shaping regional climate and hydrological patterns during winter.

As we transition into the pre-monsoon season, the balance between the westerlies and the monsoon begins to shift. The key influential zones migrate north and westward, although they exhibit lower attention weights compared to other seasons (Fig 6, B). This shift suggests that the monsoon is in its formative stages, gradually gaining strength but not yet fully exerting its influence. The movement of these zones indicates the nascent phase of monsoon development, where the system is starting to assert its presence but remains relatively subdued.

With the onset of the monsoon season, the influential zones continue their north and westward progression, now accompanied by significantly higher attention weights (Fig 6, C). This pronounced shift reflects the maturation and intensification of the monsoon, as it establishes a stronger and more extensive influence over the geographic region. The increased attention weights signify the robust circulation patterns of the monsoon, highlighting its capacity to dominate atmospheric dynamics and drive substantial climatic changes during this period.

In the post-monsoon season, there is a clear withdrawal of the key influential zones from the northwest, as they retreat towards the default winter state (Fig 6, D). This retraction signifies the diminishing strength of the monsoon, allowing the westerlies to regain their previous influence over the region. The cyclical movement of these influential zones throughout the seasons illustrates the dynamic interplay between the westerly winds and the monsoon circulation, demonstrating how their impact evolves in response to the monsoon's varying intensity and progression.

While we present seasonal spatial analysis here to highlight the westerly-monsoon interaction, HydroTrace also enables spatial analysis at the input data frequency, daily in this case, which we have illustrated in an animated image on HydroTrace Whisperer (see next section). Once again, such detailed spatial analysis is not achievable with traditional equation-driven models or existing algorithm-driven models, further showcasing the superior interpretability and effectiveness of HydroTrace in hydrological studies.

**Case Study of HydroTrace Applications in Hydropower Management**

We developed HydroTrace Whisperer, a web application that interprets attention weights generated by HydroTrace, to advance research and address the challenges faced by end-users of hydrological models, such as those managing run-of-river hydropower facilities. These facilities rely heavily on the natural flow of rivers to generate electricity, with minimal water storage. Consequently, their operation is highly sensitive to fluctuations in river streamflow, necessitating precise and timely predictions for efficient energy production. Operational bottlenecks include managing daily and seasonal variability in water flow, understanding the spatial contributions of upstream features, and planning sustainable operations under data constraints.

HydroTrace Whisperer processes attention weights extracted from HydroTrace, which is trained using the end user's streamflow data—in this case, data from a run-of-river hydropower facility near the Yangcun site—and integrates it with a language model (LLM). This integration enables users to query the hydrological system in plain language and receive intuitive visualizations and explanations, empowering facility operators to make data-driven decisions tailored to their operational needs.

A demo is available for testing at http://8.140.29.25:7860, where users can see the animation of daily spatial attention variations and ask questions such as, "How do the impacts of daily precipitation and glacier streamflow interact during 2015-06-01 to 2015-06-10?", "Which locations contribute most to the impact of glacier streamflow in March?", "How do the impacts of daily precipitation and glacier streamflow interact over the months?", and "In which month does daily precipitation play the largest role?". These queries are answered with visualizations and LLM-generated plain-language explanations, helping end users plan targeted observations and predict streamflow dynamics most relevant to their production needs.

**Discussion**

HydroTrace, which utilizes a dual attention mechanism, has demonstrated exceptional predictive accuracy and generalizability in the challenging environment of the Tibetan Plateau, an area known for its data scarcity and complex hydrological processes. HydroTrace effectively handled both in-sample and out-of-sample data. The model's consistently high performance at two distinct hydrological sites, Yangcun and Pondo, using the same algorithmic structure, underscores its robustness as a data-agnostic hydrological model.

HydroTrace outperformed traditional equation-driven models with available reports on the Yangcun and Pondo sites(*30–32*) across multiple hydrological metrics (Fig. 7). During calibration, it achieved Nash-Sutcliffe Efficiency (NSE) values of 0.98 at both Yangcun and Pondo, surpassing all existing models (Fig. 7A, 7B, and 7C). This exceptional performance highlights HydroTrace's ability to capture the complex dynamics of streamflow in a highly variable environment. During validation, it maintained strong performance with NSE values of 0.68 at Yangcun (Fig. 6A) and 0.73 at Pondo (Fig. 7,B). These values exceeded those of established equation-driven models in the region, such as HYMOD_DS at Yangcun (NSE: 0.56)(*32*) and THREW at Pondo (NSE: 0.67)(*30*). Although HydroTrace was outperformed by SIMHYD_SNOW(*31*) at Pondo during validation (SIMHYD_SNOW NSE: 0.87), its consistently high performance across both locations underscores its superiority compared to SIMHYD_SNOW, which performs better at Pondo but less effectively at Yangcun.

Hydrological modeling is one of the most representative physical process models in Earth system studies due to its interactions with the atmosphere, cryosphere, biosphere, and lithosphere(*33*). The success of HydroTrace suggests that AI-driven, data-agnostic models with dual attention mechanisms can effectively capture and explain complex spatiotemporal patterns in Earth system processes. This has considerable implications for other Earth system models, where traditional equation-driven approaches may struggle with non-linear interactions and high-dimensional data, and existing algorithm-driven approaches with interpretability. By demonstrating superior predictive accuracy and generalizability, HydroTrace paves the way for applying similar AI-driven methodologies to other components of the Earth system, such as atmospheric circulation models, ocean dynamics, and ecosystem models. Adopting such models can enhance our ability to predict and understand complex environmental phenomena, ultimately contributing to more effective management of Earth's resources and responses to global environmental challenges.

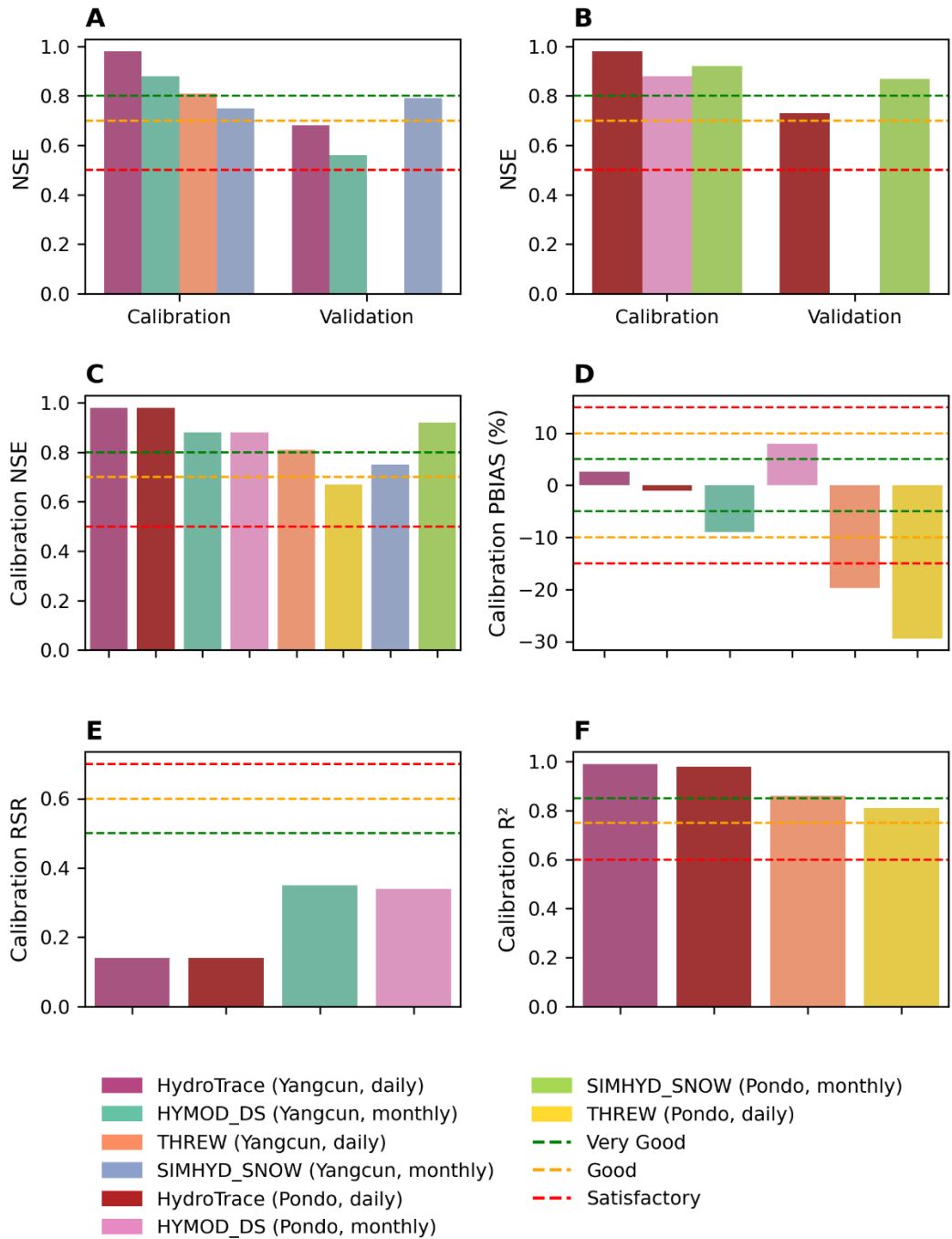

**Figure 7. Performance comparison of HydroTrace and equation-driven hydrological models**(*30–32*)**.** (A, B) NSE comparisons for calibration and validation periods at Yangcun and Pondo, respectively. (C-F) Calibration comparisons across all models for NSE, PBIAS, RSR, and R². Benchmark thresholds for streamflow estimation are shown as dashed lines: green (Very Good), orange (Good), and red (Satisfactory). Models are colored and labeled in the legend by name, hydrological site location, and streamflow data frequency.

**Advancing Interpretability in Earth System Modeling through Attention Mechanisms**

A key advantage of HydroTrace over traditional equation-based models and black-box algorithm-driven models is its interpretability. The dual attention mechanism not only improves model performance but also provides valuable insights into complex Earth system processes that are traditionally difficult to decipher. This interpretability can be viewed through three fundamental aspects, which can be extended to other domains of Earth system science.

*Revealing Interactions Among Earth System Components*
HydroTrace's attention mechanisms enable the model to uncover critical interactions among different components of the Earth system. In the context of hydrology, the model highlights how snow and glaciers contribute to streamflow, particularly during winter and pre-monsoon seasons. It captures the influence of monsoon movement on glacial melt, as well as the role of snow and surface albedo in regulating melt dynamics.

This approach can be generalized to other Earth system processes. For instance, in cryosphere-ocean interactions, attention mechanisms could help identify how sea ice melt influences ocean currents and climate patterns. By revealing these complex interdependencies, models can improve predictions of global climate dynamics and enhance our understanding of processes like thermohaline circulation, which are critical for climate modeling.

*Enhancing Spatial and Temporal Resolution of Key Processes*
By interpreting attention weights as indicators of process importance, HydroTrace quantifies and spatially resolves the contributions of various features over time. This capability allows for detailed analysis of top contributing factors by season and their spatial distributions, providing granularity not achievable with traditional models.

Extending this to other Earth system domains, attention-based models can enhance the spatial and temporal resolution of key processes. In terrestrial carbon cycling, such models could quantify the contributions of different vegetation types, soil processes, and microbial activities to carbon fluxes across various regions and seasons. This would improve our ability to predict carbon sequestration and emissions, informing climate change mitigation strategies and ecosystem management.

*Capturing Dynamic Spatial Interactions*
HydroTrace effectively captures dynamic spatial interactions by utilizing its dual attention mechanism to identify influential factors and locations over time. In hydrology, this includes understanding the interplay between westerly winds and monsoonal influences on streamflow patterns, as well as spatial variations in land surface features affecting hydrological responses. Similar methodologies could be applied to study broader spatial connections in Earth systems such as:

*Teleconnections in Climate Systems*
Attention-based models can uncover teleconnections—climate anomalies related to each other at large distances—such as the El Niño-Southern Oscillation (ENSO). By identifying and quantifying these spatial links between distant regions, models can improve predictions of global weather patterns and climate anomalies. For instance, understanding how sea surface temperature anomalies in the Pacific Ocean influence precipitation patterns in North America can enhance seasonal forecasting and climate modeling(*34*).

*Ecological Connectivity and Biodiversity*
In ecology, attention mechanisms could model spatial connections affecting species distribution and migration patterns. By highlighting critical habitats and corridors, models can inform conservation efforts and land-use planning. For example, understanding how landscape fragmentation impacts wildlife movement can aid in designing effective wildlife corridors and protecting biodiversity(*35*).

*Geological Processes and Hazard Prediction*
In geophysics, attention mechanisms could be used to model spatial connections in tectonic activity, seismic wave propagation, or volcanic activity. By identifying regions where stress accumulation might lead to earthquakes or eruptions, these models can contribute to hazard assessment and disaster preparedness(*36*).

*Oceanographic Dynamics*
Attention-based models can also be applied to oceanography to understand spatial connections in ocean currents, temperature gradients, and salinity patterns. This can improve predictions of marine heatwaves, nutrient transport, and the spread of pollutants, which are essential for marine ecosystem management and climate studies(*37*).

The interpretability capacity of HydroTrace opens up new opportunities to enhance our understanding across various Earth system domains. By using attention mechanisms, HydroTrace can reveal complex interactions, improve spatial and temporal resolution, and capture dynamic spatial dependencies—capabilities that extend beyond hydrology to other Earth system fields such as atmospheric sciences, oceanography, and ecosystem modeling. These capabilities could revolutionize how Earth system models are used in environmental management, policy decision-making, and climate modeling. By providing transparent and actionable insights into complex, high-dimensional interactions, HydroTrace and similar models could enhance the accuracy and relevance of predictions in a variety of environmental sectors.

**The Paradigm Shift in Earth System Modeling**

The transition from deterministic, physics-based models to data-agnostic machine learning frameworks is transforming how we understand and predict Earth system dynamics. Central to this shift is the use of attention mechanisms, which allow for dynamic, data-driven adaptation to complex spatiotemporal interactions within the Earth system. Proven effective across fields like natural language processing and time-series forecasting, attention mechanisms(*22*), as demonstrated by HydroTrace, are transforming Earth system models. As Earth system science moves beyond its equation-based past, the integration of AI-driven models, such as HydroTrace, exemplifies a new paradigm, ushering in the era of more interpretable, adaptable, and scalable environmental prediction and management.

*The New Era of Earth System Model Evolution*
The evolution of Earth system modeling has unfolded over several decades, with each phase marking a leap in how we conceptualize and predict interactions within the Earth system:

Pre-1980s: The early models were largely deterministic, grounded in physics-based equations, and focused on individual components like atmospheric or hydrological processes. These models were limited by computational power and simplistic assumptions, with Earth system interactions often left unexplored(*2*).

1980s–1990s: With advances in computing, Earth system models began to integrate interactions between the atmosphere, hydrosphere, and biosphere. Despite the increased complexity, these models still relied heavily on equations and struggled with non-linear interactions(*2*, *3*).

2000s: This era introduced data assimilation techniques and a push for integrated models capable of simulating dynamic Earth system processes. Still, these models continued to rely on equation-based approaches, facing challenges with high-dimensional data(*2*).

2010s: The introduction of machine learning marked a shift toward more flexible models that could handle large, complex datasets. However, many of these models were "black boxes," offering little insight into how they arrived at their predictions(*2*, *12*, *13*).

2020s: The emergence of interpretable AI4Sience models beyond traditional equation-based models and black-box algorithm-driven models like HydroTrace, marks the initiation of the evolution, which could blossom into world simulators for scientific insights and real-world applications with the state-of-the-art spatiotemporal techniques(*38–40*) in the coming decades.

*Real-World Applications: Moving from Paper to Practice*

HydroTrace's capabilities extend beyond theoretical modeling into practical, real-world applications, demonstrating its relevance and potential across various domains. In hydropower management, for example, HydroTrace provides a robust tool for predicting water flow and optimizing energy production by capturing the dynamic interplay between rainfall, snowmelt, and streamflow. The development of HydroTrace Whisperer, a web application that interprets the attention weights generated by the model, enables hydropower facility operators to make data-driven decisions. By integrating with a large language model (LLM), HydroTrace Whisperer allows users to query the hydrological system in plain language, receiving intuitive visualizations and explanations. This ease of use empowers operators to manage daily and seasonal variability in water flow and make informed decisions about sustainable operations, even in data-sparse regions.

Beyond hydropower, the model's flexibility and focus on spatiotemporal variations make it highly adaptable to other domains. In agriculture, for instance, HydroTrace could optimize irrigation schedules by analyzing the relationship between soil moisture, weather forecasts, and crop water requirements. This not only supports sustainable water use but also enhances food security by improving crop yields. Similarly, in urban planning, attention-based models could predict and manage urban heat islands by identifying key contributing factors and suggesting interventions such as green space design. This would help mitigate heat effects, reduce energy consumption, and improve urban livability. In environmental management, HydroTrace could assist in identifying critical habitats or vulnerable regions in the context of climate change, informing conservation efforts and policy decisions.

The integration of Earth system models like HydroTrace with LLMs make it possible to query complex environmental processes in real-time using natural language. This integration transforms Earth system modeling from a theoretical exercise to a powerful tool for decision-making in real-world applications.

*Unification of Earth System Modeling Through Attention*

The future of Earth system modeling lies in the unification of diverse scientific domains through attention-based algorithms. Unlike traditional, equation-driven models, attention-based frameworks like HydroTrace have the potential to seamlessly integrate multiple Earth system components—such as hydrology, climate, ecosystems, and human systems—into a cohesive model. This integration is made possible by the flexibility of attention mechanisms, which dynamically focus on relevant features and spatial locations across time, adapting to the complexities of real-world data.

These systems could function as a kind of "digital twin" for Earth—but not in the traditional sense of a replication(*41*), but rather as a data-agnostic world simulator that learns from real-time observations and adapts dynamically to ever-changing environmental conditions. Unlike conventional digital twins that rely on rigid models, this approach is more flexible, data-driven, and continuously updated, offering a more adaptable solution to Earth system modeling.

The attention-based framework offers a more dynamic, data-driven approach, making it easier to incorporate feedback loops and capture the interplay of Earth's processes in real-time. The future of Earth system science will be characterized by models that are not only predictive but also interpretable, offering a clearer understanding of how Earth's components interact and how to manage them sustainably.

**Limitations and Future Directions**

While HydroTrace shows great promise, it is important to acknowledge its limitations. The model relies on high-quality gridded land surface data, which may not be available in all regions. Efforts to improve data collection and develop methods for handling sparse or low-quality data would expand the model's applicability. Additionally, the complexity of the attention mechanisms, while powerful, may present challenges in interpretability for non-experts. Developing user-friendly interfaces and visualization tools, as demonstrated with HydroTrace Whisperer, can mitigate this issue and enhance accessibility.

Future research should focus on validating HydroTrace across diverse environments and extending the dual attention mechanism to other Earth system models. This includes exploring its application in atmospheric sciences, oceanography, and ecosystem dynamics. Enhancing computational efficiency and integrating real-time data streams could further improve the model's utility, making it suitable for operational forecasting and decision support systems.

**Materials and Methods**

**Experimental Design**

This study presents HydroTrace, an algorithm-driven hydrological modeling framework designed for complex earth systems. Unlike traditional equation-driven models, which rely on region-specific parameterizations and face challenges in adapting to diverse conditions, HydroTrace is algorithm-driven, capturing complex, nonlinear interactions among climatic, environmental, and hydrological factors. By integrating spatiotemporal and feature-wise attention mechanisms, HydroTrace ensures universal applicability, delivering consistent predictive accuracy and interpretability across varied scenarios without requiring custom structural design.

**Setup**

We used Python 3.8 on a Linux system equipped with a 32-core CPU, 188 GB of memory, and an A10 GPU for data processing and modeling in this study.

**Data Preparation**

*Data Sources*

Streamflow Data: Licensed daily streamflow measurements, together with the coordinates of the measurement site are provided by the hydropower facilities operating near the Yangcun and Pondo sites.

Environmental and Climatic Data: Spatially gridded datasets encompassing glaciers(*42*) and other hydrological variables(*43*) are retrieved from the National Snow and Ice Data Center (NSIDC) at the University of Colorado Boulder (CU Boulder). Locally calibrated daily precipitation(*44*), elevation and land cover data(*45*) are retrieved from National Tibetan Plateau Data Center (TPDC).

*Preprocessing*

Preprocessing focused on creating temporally and spatially structured input data for model training. Key steps included:

Cyclic Encoding of Spatial Coordinates: Latitude and longitude were cyclically encoded to capture their periodic nature:

$$\sin\_\text{lat} = \sin\left(\frac{2\pi \cdot \text{lat}}{90}\right), \quad \cos\_\text{lat} = \cos\left(\frac{2\pi \cdot \text{lat}}{90}\right)$$

$$\sin\_\text{lon} = \sin\left(\frac{2\pi \cdot \text{lon}}{180}\right), \quad \cos\_\text{lon} = \cos\left(\frac{2\pi \cdot \text{lon}}{180}\right)$$

Sliding Windows: Temporal sequences of seven days (window size= 7) were applied to provide context for streamflow predictions.

Let $X_t \in \mathbb{R}^{H \times W \times C}$ represent the spatial data at time $t$, where

- $H$: Height of the grid (number of latitude points).
- $W$: Width of the grid (number of longitude points).
- $C$: Number of features per grid cell.

The input sequence for a window size *W* is:

$$\mathcal{X}_i = \{X_t, X_{t+1}, \ldots, X_{t+W-1}\}, \quad t = i, \ldots, i+W-1$$

with corresponding output:

$$\mathbf{y}_i = x_{i+W}$$

Static Features Integration: Elevation and land cover data were added as static features, repeated along the temporal axis to match dynamic inputs.

**HydroTrace Model**

*Model Architecture*

The HydroTrace model integrates its customized attention mechanisms(*22*) with customized ConvLSTM(*46*) layers to capture spatiotemporal and feature-wise dynamics. The architecture consists of:

- Customized ConvLSTM Layers: Extract spatiotemporal features by processing multichannel inputs.

- Attention Mechanisms: Provide interpretability and focus on the most influential spatial regions and features.

*Customized ConvLSTM Layer*

**The Depthwise ConvLSTM2D layer** processes each feature channel separately using ConvLSTM2D, capturing spatiotemporal dependencies within each channel without mixing information across channels.

For each feature channel $c \in \{1, 2, \ldots, C\}$ and each timestep $t \in \{1, 2, \ldots, T\}$, compute:

Input gate: $\mathbf{i}_t^c = \sigma\left(\mathbf{W}_{xi}^c * \mathbf{X}_t^c + \mathbf{W}_{hi}^c * \mathbf{h}_{t-1}^c + \mathbf{b}_i^c\right)$

Forget Gate: $\mathbf{f}_t^c = \sigma\left(\mathbf{W}_{xf}^c * \mathbf{X}_t^c + \mathbf{W}_{hf}^c * \mathbf{h}_{t-1}^c + \mathbf{b}_f^c\right)$

Cell Candidate: $\tilde{\mathbf{c}}_t^c = \tanh\left(\mathbf{W}_{xc}^c * \mathbf{X}_t^c + \mathbf{W}_{hc}^c * \mathbf{h}_{t-1}^c + \mathbf{b}_c^c\right)$

Output Gate: $\mathbf{o}_t^c = \sigma\left(\mathbf{W}_{xo}^c * \mathbf{X}_t^c + \mathbf{W}_{ho}^c * \mathbf{h}_{t-1}^c + \mathbf{b}_o^c\right)$

Cell State Update: $\mathbf{c}_t^c = \mathbf{f}_t^c \odot \mathbf{c}_{t-1}^c + \mathbf{i}_t^c \odot \tilde{\mathbf{c}}_t^c$

Hidden State Update: $\mathbf{h}_t^c = \mathbf{o}_t^c \odot \tanh\left(\mathbf{c}_t^c\right)$

where:

- $\sigma$ is the sigmoid activation function.
- $*$ denotes convolution operation.
- $\odot$ denotes element-wise multiplication.
- $\mathbf{W}$ and $\mathbf{b}$ are the convolutional weights and biases specific to channel $c$.

The output of the layer concatenates the outputs from all channels along the feature dimension:

$$\mathbf{H}_t = \text{Concat}\left(\mathbf{h}_t^1, \mathbf{h}_t^2, \ldots, \mathbf{h}_t^C\right) \in \mathbb{R}^{B \times T \times H \times W \times C}$$

*Attention Mechanisms*

HydroTrace leverages two attention mechanisms:

***TimedistributedSpatial Attention Weights*** highlight the spatial locations most relevant to the model's predictions. These weights are computed using convolutional layers and a softmax activation:

$$\alpha_{i,j} = \sigma(\text{Conv2D}(x_{i,j}))$$

where :
- $\alpha_{i,j}$ : Spatial attention weight for grid point *(i,j)*,
- $\mathbf{x}_{i,j}$ : Input data at grid point *(i,j)*,
- The $\sigma()$ activation function is applied to normalize the attention weights. For an input vector $z = [z_1, z_2, \ldots, z_n]$, the function normalize attention weights by mapping the values to a range between 0 and 1:

$$\sigma(z_i) = \frac{1}{1 + e^{-z_i}}$$

- Conv2D: Convolutional layer applied to the input. The Conv2D operation is a building block in convolutional neural networks, used to extract spatial features from input data. Mathematically, the output of a Conv2D operation for an input x and a kernel k at a specific position (i,j) is given by:

$$y(i, j) = \sum_{m=1}^{M} \sum_{n=1}^{N} x(i+m-1, j+n-1) \cdot k(m,n)$$

where:
  x(i, j) is the input value at grid point (i,j),
  k(m,n) is the kernel value at position (m,n),
  M and N are the height and width of the kernel,
  y(i,j) is the output value at grid point (i,j).

***Feature-Wise Attention Weights*** evaluate the importance of each feature in the input dataset for streamflow prediction. These weights are computed as:

$$\beta_k = \text{softmax}(\mathbf{W}_k \cdot \mathbf{x})$$

where:
- $\mathbf{W}_k$ is a learnable weight matrix for feature k,
- $\mathbf{x}$ is the input feature vector.
- The softmax activation function is applied to normalize the attention weights. For an input vector $z = [z_1, z_2, \ldots, z_n]$, the function outputs a probability distribution such that the sum of all probabilities equals 1:

- $$\mathrm{softmax}(z_i) = \frac{e^{z_i}}{\sum_{j=1}^{n} e^{z_j}}$$

Feature-wise attention identifies the relative contribution of input variables, providing insights into their role in estimation.

The attention mechanisms are trained end-to-end within the model, and their outputs are saved for post-analysis.

*Attention Weights Processing*

Attention weights were extracted during model inference for spatial and feature interpretability and preprocessed as follows:

$$\mathbf{X}^{\mathrm{att}} = \alpha \odot \beta \odot \mathbf{X}$$

where:

- $\alpha$ is the spatial attention weight matrix,
- $\beta$ is the feature-wise attention weight vector,
- $\mathbf{X}$ is the original input tensor,
- $\odot$ denotes element-wise multiplication.

Mapping to Geographic Coordinates: Attention weights were linked to latitudinal and longitudinal grids for spatial analysis.

Temporal Aggregation: Weights were aggregated over time to highlight persistent patterns.

*Training and Validation*

The HydroTrace model was trained using a sliding-window approach, where sequences of seven consecutive days were used as input, and the eighth day served as the target streamflow value. To ensure a robust evaluation, the dataset was split into 80% for training and 20% for validation, with the split designed to maintain temporal diversity. This means the validation data was not concentrated in a single season but rather distributed across the dataset's full temporal range, enabling the model's performance to be evaluated under diverse climatic and hydrological conditions. The model was trained using a batch size of 8, and hyperparameters were tuned using Keras Tuner's RandomSearch algorithm over 10 trials. The best model was selected based on the validation mean absolute error (MAE). During evaluation, predictions were generated for the validation dataset, and the statistical metrics were computed to assess model performance.

During training, the following strategies were implemented to optimize the model and prevent overfitting:

Early Stopping: Training was halted when the validation performance plateaued, ensuring computational efficiency and mitigating overfitting.

Learning Rate Reduction: The learning rate was dynamically reduced when the validation loss showed no improvement, allowing for finer adjustments during later training stages.

## HydroTrace Whisperer App

We developed a web application using Gradio(*47*), PostgreSQL(*48*) together with Large Language Model Qwen-plus(*49*) API to deliver HydroTrace Interpretation to users with natural language interaction.

## Statistical Analysis

We evaluated the performance of HydroTrace using several common statistical metrics(*27*) to assess the accuracy and reliability of hydrological predictions. The statistical analysis was conducted using Python (version 3.8) and TensorFlow (version 2.12.1). All computations were performed using built-in functions and custom scripts, and the evaluation metrics were computed as follows.

### Data Preprocessing for HydroTrace Evaluation

The validation dataset consisted of *N* samples, each containing time-series data with dimensions corresponding to *T* (time window size) ×*W* (longitude units) ×*H* (latitude units) ×*C* (feature dimension). Missing values *NaN* in the input data were handled by computing the mean of the non-*NaN* values and replacing any *NaN* values with this mean:

$$X_{\text{mean}} = \frac{1}{n_{\text{non-NaN}}} \sum_{i \in \text{non-NaN}} X_i \quad X_{\text{filled}} = \begin{cases} X_i & \text{if } X_i \text{ is not NaN,} \\ X_{\text{mean}} & \text{if } X_i \text{ is NaN.} \end{cases}$$

Rows in the output data that contained *NaN*s were removed, along with the corresponding rows in input data, resulting in a final dataset of *N′* samples for analysis. In this study, *N′*=356 for Yangcun and *N′*=359 for Pondo. For both Yangcun and Pondo, *W*=79, *H*=292, *C*=49.

### Model Evaluation Metrics

The performance of the model was evaluated using the following statistical metrics:

### Nash-Sutcliffe Efficiency (NSE)

The Nash-Sutcliffe Efficiency(*26*, *27*, *50*) measures how well the predicted values match the observed data, with a value of 1 indicating a perfect match and a value of 0 indicating that the model predictions are as accurate as the mean of the observed data. The NSE in this study is calculated as:

$$\text{NSE} = 1 - \frac{\sum_{i=1}^{n}(O_i - P_i)^2}{\sum_{i=1}^{n}(O_i - \bar{O})^2}$$

where:

- $O_i$ : Observed value at index $i$ ,
- $P_i$ : Predicted value at index $i$ ,
- $\bar{O}$ : Mean of the observed values ,
- $n$ : Number of samples .

### *Percent Bias (PBIAS)*

Percent Bias(*26*, *27*) measures the average tendency of the predicted values to be larger or smaller than their observed counterparts. It is expressed as a percentage:

$$\text{PBIAS} = 100 \times \frac{\sum_{i=1}^{n}(P_i - O_i)}{\sum_{i=1}^{n} O_i}$$

### *RMSE-Observations Standard Deviation Ratio (RSR)*

The RMSE-Observations Standard Deviation Ratio(*26*) standardizes the root mean square error (RMSE) using the standard deviation of the observations. It is calculated as:

$$\text{RSR} = \frac{\text{RMSE}}{S_{obs}} = \frac{\sqrt{\frac{1}{n}\sum_{i=1}^{n}(P_i - O_i)^2}}{\sqrt{\frac{1}{n-1}\sum_{i=1}^{n}(O_i - \bar{O})^2}}$$

### *Coefficient of determination ($R^2$)*

Coefficient of determination (*$R^2$*) (*26*, *27*) is a statistical measure that represents the proportion of the variance in the dependent variable that is predictable from the independent variables. It provides an indication of the goodness of fit of the model, with values ranging from 0 to 1. An *$R^2$* value of 1 indicates perfect prediction, while an *$R^2$* value of 0 indicates that the model does not explain any of the variance in the data.

In this study, *$R^2$* was computed explicitly using the Pearson correlation coefficient (**r**) between the observed and predicted values, as follows:

$$r = \frac{\text{Cov}(O,P)}{\sigma_O \sigma_P},$$

$$R^2 = r^2$$

**Acknowledgments**

**Funding:**

Xizang Science and Technology Plan Project, grant number XZ202403ZY0018 (CX,YL,DC,YL,AX).

National Natural Science Foundation of China, grant number 42250610213(DK).

Swedish Research Council (VR), grant numbers 2021-02163 and 2022-06011 (DC).

**Author contributions:**
Conceptualization: CX, LY, DC
Methodology: CX, LY, DC, YL
Data curation: LY, CX
Investigation: LY, AX, ZL, QH, CX, GZ
Software: CX, YL, LL, MZ
Visualization: CX, YL, GZ
Supervision: DC
Writing—original draft: CX, LY, HY
Writing—review & editing: DC, AX, ZL, QH, GZ, DK
Project administration: WW, CX

**Competing interests:** Authors declare that they have no competing interests.

**Data and materials availability:** All data, materials, and methods used in the analyses are available for the purposes of reproducing or extending the findings, with some restrictions: custom scripts and algorithmic implementation details are currently undergoing the patent application process. The HydroTrace model can be accessed through an API to enable reproduction and further analysis upon publication. Licensed daily streamflow measurements and the coordinates of the measurement sites are provided by the hydropower facilities operating near the Yangcun and Pondo sites. This data is not publicly available and requires prior consent from the data provider for access due to licensing restrictions. However, these data will be embedded in the HydroTrace API provided for reproduction purposes. Attention weights from the HydroTrace model are accessible via the HydroTrace Whisperer app (http://8.140.29.25:7860) to support extension of the analyses.